\documentclass{article}

\usepackage{PRIMEarxiv}

\usepackage[utf8]{inputenc} % allow utf-8 input
\usepackage[T1]{fontenc}    % use 8-bit T1 fonts
\usepackage{hyperref}       % hyperlinks
\usepackage{url}            % simple URL typesetting
\usepackage{booktabs}       % professional-quality tables
\usepackage{amsfonts}       % blackboard math symbols
\usepackage{nicefrac}       % compact symbols for 1/2, etc.
\usepackage{microtype}      % microtypography
\usepackage{lipsum}
\usepackage{fancyhdr}       % header
\usepackage{graphicx}       % graphics
\graphicspath{{media/}}     % organize your images and other figures under media/ folder

\usepackage{amsmath}
\usepackage{graphicx}
\usepackage{multirow}
\usepackage{caption}
\usepackage{subcaption}

\title{Multimodal Quasi-AutoRegression: Forecasting the visual popularity of new fashion products

}

\author{
  Stefanos I. Papadopoulos, 
  Christos Koutlis,
  Symeon Papadopoulos, 
  Ioannis Kompatsiaris
  \\
  \texttt{stefpapad@iti.gr, ckoutlis@iti.gr, papadop@iti.gr, ikom@iti.gr} \\
  CERTH-ITI, Greece\\
}

\begin{document}
\maketitle

\begin{abstract}

Estimating the preferences of consumers is of utmost importance for the fashion industry as appropriately leveraging this information can be beneficial in terms of profit. Trend detection in fashion is a  challenging task due to the fast pace of change in the fashion industry. 
Moreover, forecasting the visual popularity of new garment designs is even more demanding due to lack of historical data.
To this end, we propose MuQAR, a Multimodal Quasi-AutoRegressive deep learning architecture that combines two modules: (1) a multi-modal multi-layer perceptron processing categorical, visual and textual features of the product and (2) a quasi-autoregressive neural network modelling the ``target'' time series of the product's attributes along with the ``exogenous'' time series of all other attributes. 
We utilize computer vision, image classification and image captioning, for automatically extracting visual features and textual descriptions from the images of new products.
Product design in fashion is initially expressed visually and these features represent the products' unique characteristics without interfering with the creative process of its designers by requiring additional inputs (e.g manually written texts).
We employ the product's target attributes time series as a proxy of temporal popularity patterns, mitigating the lack of historical data, while exogenous time series help capture trends among interrelated attributes.
We perform an extensive ablation analysis on two large scale image fashion datasets, Mallzee and SHIFT15m to assess the adequacy of MuQAR and also use the Amazon Reviews: Home and Kitchen dataset to assess generalisability to other domains. A comparative study on the VISUELLE dataset, shows that MuQAR is capable of competing and surpassing the domain's current state of the art by 4.65\% and 4.8\% in terms of WAPE and MAE respectively.

\end{abstract}

% keywords can be removed

\section{Introduction}

Fashion is a dynamic domain, and fashion trends and styles are highly time-dependent. Systematic analysis of fashion trends is not only useful for consumers who want to be up-to-date with current trends, but is also vital for fashion designers, retailers and brands in order to optimize production cycles and design products that customers will find appealing when they hit the shelves. 
Moreover, it could potentially help mitigate the problem of unsold inventory in fashion which is caused by a mismatch between supply and demand \cite{ekambaram2020attention} and has a significant environmental impact;  with million tonnes of garments ending up in landfills or being burned every year \cite{niinimaki2020environmental}.

At the same time, fashion is a primarily visually-driven domain.
As a result, computer vision has successfully been utilized to assist fashion recommendations and trend forecasting \cite{cheng2021fashion}.
Recent studies have utilized visual features - extracted by computer vision models - in order to identify fashion styles \cite{al2017fashion} or attributes \cite{mall2019geostyle} and then detect and analyse trends in fashion. 
However, such approaches are limited to detecting coarse-level trends and can not work for specific garment designs. 
They can forecast whether ``chunky trainers'' will be trending this season but all ``chunky trainers'' will receive the same popularity score. Specific visual differences in individual garments are not taken into consideration.  
Autoregressive (AR) neural networks have been used for forecasting the popularity of specific garments based on their past popularity \cite{lo2019dressing}. However, new products by definition lack historical data which renders the use of conventional AR networks impracticable. 
Few recent research works have addressed sales forecasting of new garments, by utilizing KNN-based (nearest neighbors) \cite{craparotta2019siamese}, auto-regressive networks with auxiliary features (images, fashion attributes and events) \cite{ekambaram2020attention} or a non-AR Transformer modelling images and fashion attributes along with the ``target'' time series of those attributes collected from Google Trends\footnote{\url{https://trends.google.com}} \cite{skenderi2021well}. 
However, fashion attributes are not always independent of each other. Trends in certain attributes may affect other interdependent attributes. If for example ``warm minimalism'' was trending in fashion, a series of light, neutral and pastel colors would show an increase in popularity while bold graphics and patterns would decrease.

\begin{figure}[ht]
    \centering
    \includegraphics[width=0.5\textwidth]{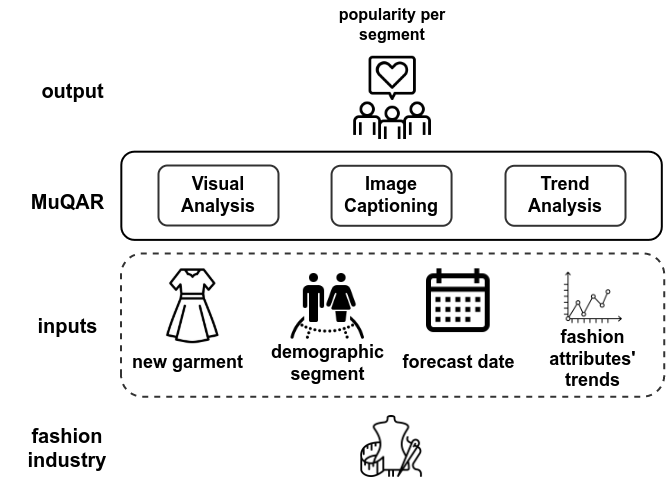}
    \caption{High level workflow of MuQAR. The new garment's image and the trends of fashion attributes are analysed by the modules of MuQAR which predicts the garment's popularity for a given demographic segment and target date.}
    \label{fig:highlevel}
\end{figure}

The objective of this study is to accurately forecast the visual popularity of new garment designs that lack historical data. 
To this end, we propose MuQAR, a Multi-Modal Quasi-AutoRegressive neural network architecture and in Figure~\ref{fig:highlevel} we illustrate its high level workflow. 
MuQAR combines two modules: (1) a multi-modal multilayer perceptron (FusionMLP) representing the visual, textual, categorical and temporal aspects of a garment and (2) a quasi-autoregressive (QAR) neural network modelling the time series of the garment's fashion attributes (target time series) along with the time series of all other fashion attributes (exogenous time series).
We expand the QAR framework by employing the logic of nonlinear autoregressive network with exogenous inputs (NARX).
Our rationale is that modelling the target time series of the garment's attributes will work as an informative proxy of temporal patterns mitigating the lack of historical data while exogenous time series will help the model identify relations among fashion attributes.

The aim of this study is to provide fashion designers with real-time feedback for their new designs without interfering in their creative process. 
In fashion, design usually begins with sketching and visual prototyping - in 2D or 3D programs - expressing the silhouette, fitment, colors and fabrics of the new garment. 
Previous works have relied on computer vision to extract relevant visual features from the garment's images along with its fashion attributes in order to forecast its popularity without interference in the creative process (e.g by requiring textual descriptions of the garment).
We expand upon this idea by utilizing image captioning (IC) for automatically extracting textual descriptions of the new garment which could provide richer descriptions and useful contextual information about the attributes of the garment. 
For example, while an attribute detection model may recognise that a ``varsity college jacket'' has a graphic design, it is black and green and is made out of leather and jersey, an IC model could also describe the position of the graphic design (e.g across the chest) colors and fabrics (e.g black leather sleeves, green jersey body).

The main contributions of our work are:

\begin{itemize}

\item Propose a novel deep learning architecture that employs the logic of NARX models in QAR for forecasting the popularity of new products that lack historical data. We compare various QAR models, including: CNN, LSTM, ConvLSTM, Feedback-LSTM, Transformers and DA-RNN.

\item Integrate image captioning in the multi-modal module for capturing contextual and positional relations of the products' attributes.

\item Collect a new large-scale fashion dataset that includes popularity scores in relation to demographic groups allowing specialised forecasts for different market segments.

\item Perform an extensive ablation study on three datasets (two fashion and one home decoration) to assess the validity and generalisability of the proposed methodology. A comparative study on a fourth fashion-related dataset shows that our model surpasses the domain's state of the art by 4.65\% and 4.8\% improvements in terms of WAPE and MAE respectively.

\end{itemize}

\begin{figure*}[!ht]
    \centering
    \includegraphics[width=\textwidth]{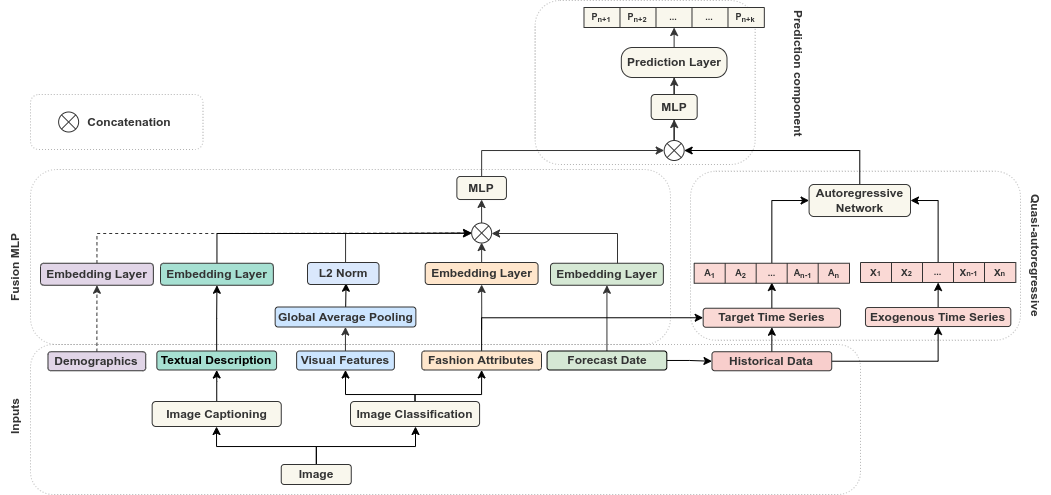}
    \caption{Architecture of MuQAR. Intermittent arrows are optional 
    % demographics embedding layer is optional 
    and applicable only to the Mallzee dataset.}
    \label{fig:MuQAR}
\end{figure*}

\section{Related Work}
\label{sec:RW}

Deep learning has been used for time series forecasting in numerous domains including climate modelling, biological sciences, medicine \cite{lim2021time}, music, meteorology, solar activity, finance \cite{koutlis2020lavarnet} and among other industries, in fashion \cite{chang2021fashion}.
Additionally, researchers have been experimenting in recent years with the inclusion of visual information for detecting fashion trends and forecasting the popularity of garments or complete outfits \cite{cheng2021fashion}. 

Visual features extracted by computer vision models have been used to discover fashion styles \cite{al2017fashion} or fashion attributes \cite{mall2019geostyle} and neural networks to detect fashion trends.
More recent works have examined how fashion trends from one city may influence other cities \cite{al2020paris} or how knowledge enhanced neural networks can improve fashion trend detection \cite{ma2020knowledge}. 
The aforementioned studies extracted fashion styles or attributes from fashion imagery in order to forecast fashion trends. A significant limitation of these approaches however is that they can only work on a coarse level but not a finer level.
Trends may show that “floral dresses will be trending next spring” but all new “floral dress” designs will receive the same score and can not produce informed and specialized predictions for individual garments or outfits. Autoregressive (AR) networks have also been used to forecast the visual popularity for specific garments or outfits \cite{lo2019dressing}. 
Nevertheless, new garment design, by definition, lack historical data and therefore conventional AR networks - that rely on past sequences to predict future outcomes - can not be utilized.

While no works have directly addressed popularity forecasting of new garments few recent research works have addressed sales forecasting of new garments, which are closely related.
Loureiro et. al. \cite{loureiro2018exploring} and Singh et. al. \cite{singh2019fashion} were two of the first works to utilize regressive deep learning for sales forecasting of new garments but did not utilize visual features. 
Craparotta et. al. \cite{craparotta2019siamese} proposed a KNN-based approach relying on an image similarity network connected to a siamese network for forecasting sales of new garments. 
Ekambaram et. al. \cite{ekambaram2020attention} utilized an AR multi-modal RNN that combines visual and textual features with exogenous factors (holidays, discount season, events etc). Since new products do not have historical data, the authors prepend two starting delimiters and utilize teacher enforcing for the proceeding steps during the training process. 
Skenderi et. al. \cite{skenderi2021well} criticised the reliance on purely AR networks for new product sale forecasting because of the compounding effect caused by first-step errors. 
Instead, they propose GTM-Transformer, a multi-modal, non-AR Transformer that utilizes images, text and time series of the garment's attributes collected from Google Trends. 
Their work can be considered the first to utilize a version of quasi-autoregression using the ``target'' attributes time series and it was capable of outperforming both KNN-based approaches \cite{craparotta2019siamese} and multi-modal AR \cite{ekambaram2020attention}.

However, fashion attributes are not always independent of each other. Trends in some attributes may positively or negatively affect others, e.g. complementary colors or matching categories. 
To this end, we employ the logic of nonlinear autoregressive network with exogenous inputs (NARX) \cite{billings2013nonlinear} within QAR by integrating the ``exogenous'' fashion attributes along with the ``target'' attributes of a new garment in order to forecast its popularity.

Apart from the visual features of the garment,
both \cite{ekambaram2020attention} and \cite{skenderi2021well} used fashion attributes as textual information. 
Fashion attributes can be extracted from images by specialised classification models without requiring manual annotation from the fashion designers and can offer valuable information to the overall neural network. 
The advantage of such approaches is that, in real world scenario, a forecasting model utilizing information about attributes would not interfere with the creative process of the designers.
We expand upon this idea by utilizing an image captioning model (IC) for automatically extracting full textual descriptions from the images of new garment designs. Our hypothesis is that IC could create richer descriptions that capture useful contextual and positional information about the garment's attributes that will help improve the performance of multi-modal forecasting models.

\section{Methodology}
\label{section:MuQAR}

In this study we attempt to forecast the visual popularity of new garment designs. Conventional autoregressive (AR) forecasting models can not be utilized since new products lack historical data. On the other hand, conventional regression models are not as well equipped to detect temporal trends which play an important role in the domain of fashion. 
To this end, inspired by previous works, we propose MuQAR, a multimodal quasi-autoregressive neural architecture that consists of two modules: FusionMLP and QAR as well as a final prediction component that combines them. In Figure~\ref{fig:MuQAR}, the details of its architecture are illustrated.

\subsection{FusionMLP}
The first module consists of a multimodal multilayer perceptron that processes the visual, textual, categorical and temporal features of a product. The visual features vector $F_v\in\mathbb{R}^V$, is extracted by the last convolutional layer of CNN-based networks. 
% \footnote{
As will be discussed in Section \ref{section:datasets}, we use features extracted from pre-trained networks on ImageNet for SHIFT15m, Amazon Reviews and VISUELLE to ensure comparability. However, we utilize a hierarchical deep learning network, fine tuned on fashion imagery for the Mallzee image dataset \cite{papadopoulos2022attentive}.
% }. 
After the extraction, global average pooling and L2 normalisation is applied. 

We utilize OFA \cite{wang2022unifying} - a state-of-the-art IC model on COCO Captions\footnote{\url{https://paperswithcode.com/sota/image-captioning-on-coco-captions}} - for automatically extracting textual descriptions from fashion imagery. OFA does not specialise on fashion imagery but it has been trained on a large-scale e-commerce dataset that also included numerous fashion products. We manually examined hundreds of inference texts and deemed its predictions to be very accurate. 
We pre-process the extracted captions by lower-casing, removing punctuation and stop-words and then tokenizing them. 
We use a word embedding layer that produces the $F_w\in\mathbb{R}^W$ vector based on one integer index per token in relation to the whole vocabulary W.
We apply IC on the Mallzee and VISUELLE datasets since they provide the garment's images but not on SHIFT15m and Amazon Reviews that only provide pre-computed visual features.

The catecorical features vector $F_c\in\mathbb{R}^{c_p\cdot d_c}$, is the concatenation of $c_p$ learnable embeddings of size $d_c$ each corresponding to a fashion label assigned to product $p$, defined by:
\begin{equation}
F_c = [f_1\text{E}_c;f_2\text{E}_c;\dots;f_{c_p}\text{E}_c]
\end{equation}
where $[;]$ denotes concatenation, $E_c\in\mathbb{R}^{C\times d_c}$ is the embedding matrix for fashion labels, $C$ is the total number of fashion labels and $f_i\in\mathbb{R}^C$ are one-hot encoding vectors with 1 at the index of the corresponding fashion label and zero elsewhere. The temporal feature vector $F_t\in\mathbb{R}^{4\times d_t}$, is the concatenation of 4 learnable embeddings of size $d_t$ corresponding to the day, week, month and season of the target date, defined by:
\begin{equation}
F_t = [d\text{E}_d;w\text{E}_w;m\text{E}_m;s\text{E}_s]
\end{equation}
where $E_d\in\mathbb{R}^{366\times d_t}$, $E_w\in\mathbb{R}^{52\times d_t}$, $E_m\in\mathbb{R}^{12\times d_t}$, $E_s\in\mathbb{R}^{4\times d_t}$ are embedding matrices for day, week, month and season of the year respectively and $d\in\mathbb{R}^{366}$ (leap year provision), $w\in\mathbb{R}^{52}$, $m\in\mathbb{R}^{12}$, $s\in\mathbb{R}^4$ are the corresponding one-hot encoding vectors. For the demographic group input, that is optional and considered only in one dataset here, we also consider a learnable embedding $F_g=g\text{E}_g\in\mathbb{R}^{d_g}$, accordingly. Finally, a standard MLP network with $n_{mlp}$ dense layers of $u_{mlp}$ relu activated units processes the concatenation of all features resulting in $F_{F}\in\mathbb{R}^{f}$:
\begin{equation}
F_{F} = \text{MLP}([F_v;F_w;F_c;F_t;F_g])
\end{equation}

\subsection{QAR}
The second module utilizes the product’s attributes' time series (``target'') $\{A_t\}$ along with all other attributes time series (``exogenous'') $\{X_t\}$ as input in order to predict the product's popularity time series $\{P_t\}$. 
More precisely, we feed QAR module with two matrices $\textbf{A}=\{A_1,\dots,A_n\}\in\mathbb{R}^{n\times c_p}$ 
that contains $n$ time steps prior to the forecast date for $c_p$ target fashion labels assigned to product $p$ and $\textbf{X}=\{X_1,\dots,X_n\}\in\mathbb{R}^{n\times c_x}$ for all ``exogenous'' fashion labels $c_x$. 
$X$ includes the time series for all available fashion attributes within the time period, but we set the target attributes of $c_p$ to zero in $c_x$ to avoid information leakage.

The proposed methodology, MuQAR, is a modular architecture meaning that it can integrate any AR network to the QAR module. This allows for identifying the optimal AR network for a given task. In this study we experimented with multiple AR architectures that have been used for time series forecasting, namely: 
 Long Short Term Memory network (LSTM) \cite{hochreiter1997long}, 
 (2) Feedback LSTM (F-LSTM) \cite{graves2013generating}, (3) Convolutional Neural Network (CNN) \cite{zhao2017convolutional}, (4) Convolutional LSTM (ConvLSTM) \cite{xingjian2015convolutional}, (5) Transformer \cite{vaswani2017attention} 
for experiments that only utilize $\{A_t\}$ 
and (6) Dual-Stage Attention-Based Recurrent Neural Network (DA~-~RNN) \cite{qin2017dual} adapted for multivariate time series, (7) Convolutional LSTM with exogenous time series (ConvLSTM~+~X) inspired by the encoder proposed in \cite{chang2018memory} for experiments that utilize both  $\{A_t\}$ and $\{X_t\}$. 
DA-RNN is a dual-stage architecture that first processes $\{X_t\}$ and then $\{A_t\}$ while ConvLSTM+X processes $\{A_t\}$ and $\{X_t\}$ in parallel - with two separate ConvLSTM neural networks - and then concatenates the resulting representation vectors.
After processing the input time series, QAR produces a vector representation $F_Q\in\mathbb{R}^{q}$ pertinent to the forecast.

\subsection{Prediction component}
The concatenated vector $F=[F_F;F_Q]\in\mathbb{R}^{f+q}$ is further processed by another dense layer on top of which a linear layer forecasts the product's next $k$ popularity time steps $\{P_{n+1},\dots,P_{n+k}\}$, as can be seen in Figure \ref{fig:MuQAR}. 

\section{Experimental Setup}
\label{section:experimental_setup}
\subsection{Evaluation Protocol}
\label{section:evaluation}
In order to correctly assess the task of forecasting the popularity of new products, we propose an evaluation protocol where the models are trained on \textit{established products} and evaluated on \textit{new products}. We consider as established those products that have multiple records in the dataset in different times while new are considered products that make only a single appearance in the dataset and the model has not previously encountered. We therefore split the data into established products which are used as the training set and new products which are split in half for the validation and testing sets. We apply this protocol on Mallzee, SHIFT15m and Amazon Reviews datasets but for VISUELLE we follow the experimental protocol described in \cite{skenderi2021well}. 

For the evaluation we used multiple evaluation metrics. For regression tasks we used the: Mean Absolute Error (MAE), Pearson Correlation Coefficient (PCC) and Binary Accuracy (BA) while for classification tasks we used: Accuracy and the Area under the ROC Curve (AUC). We selected the best performances of each model with the use of TOPSIS, a multi-criteria decision analysis method \cite{hwang1981methods}. 

We perform an extensive grid-search for tuning the hyper parameters of FusionMLP and the QAR networks. 
We integrate the best performing QAR network with FusionMLP's best performing hyper-parameter combinations on each dataset separately to create MuQAR.
QAR models are trained with weekly aggregated time series using 12 weeks as input and 1 week as output.

\subsection{Datasets}
\label{section:datasets}

\subsubsection{\textbf{VISUELLE}}

% \begin{table}
%   \centering
%     \caption{Details for the popularity datasets.}
%   \begin{tabular}{lll}
%     \toprule
%     \textbf{Dataset} & \textbf{Records} & \textbf{Period} 
%     \\ 
%     \midrule
%     VISUELLE & 5,577 & 2016 - 2019 
%     \\
%     SHIFT15m & 15,218,721 & 2010 - 2020 
%     \\
%     Mallzee & 5,412,193 & 2017 - 2020  
%     \\
%     Amazon Reviews &  3,002,786 & 2009 - 2014 
%     \\
%     \bottomrule
%   \end{tabular}
%   \label{tab:data_stats_pop}
% \end{table}

VISUELLE\footnote{\url{https://github.com/HumaticsLAB/GTM-Transformer}} is a public fashion image dataset that contains 12 week long sales sequences for 5577 garments spanning from October 2016 to December 2019 \cite{skenderi2021well}. For each garment, it provides an image, textual information (categorical labels related to fashion categories, fabrics, colors) and time series related to the categorical labels collected from Google Trends. 
The dataset is sorted by date and split into 5080 garments for training and 497 for evaluation. 

\begin{figure*}[ht]
    \centering
    \includegraphics[width=0.8\textwidth]{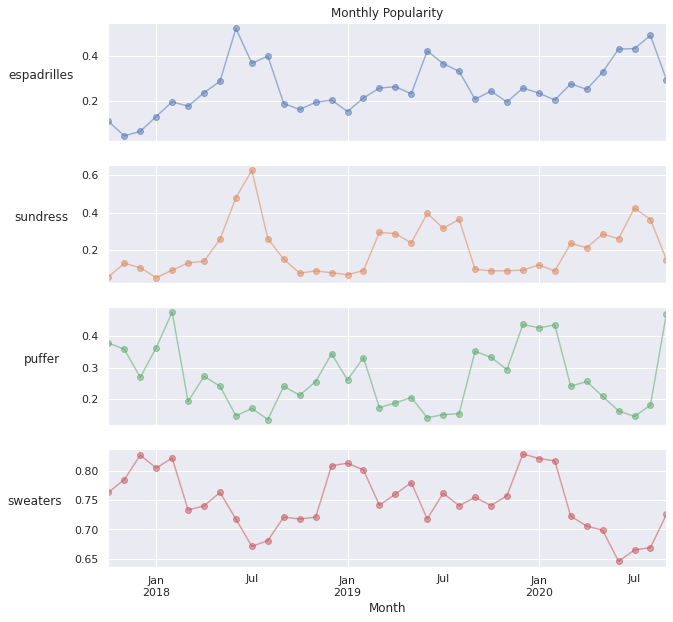}
    \caption{Monthly aggregated time series for fashion categories from the Mallzee dataset.}
    \label{fig:timeseries}
\end{figure*}

\subsubsection{\textbf{SHIFT15m}}

Due to the fact that VISUELLE is a relatively small scale dataset we also experiment with larger-scale image fashion datasets.
SHIFT15m\footnote{\url{https://github.com/st-tech/zozo-shift15m/tree/main/benchmarks}} is a public, large scale and multi-objective fashion dataset that consists of 15,218,721 outfits posted by users in a social network platform among 193,574 users and 2,335,598 garments between 2010 and 2020 \cite{kimura2021shift15m}. The dataset provides the user ID, the number of likes that the outfit has received, the date it was published and the items that constitute the outfit, including the item IDs and two types of fashion categories (comprising 7 and 43 unique categories respectively). The outfits' images are not available but SHIFT15m provides the visual features extracted by a VGG network pre-trained on the ImageNet dataset.

We re-purpose SHIFT15m for new garment popularity forecasting. We remap the initial outfit-level onto the garment-level by splitting each outfit into the individual garments that make it up and by defining the number of likes as the target variable. We assume that each garment has an equal contribution to the overall popularity of the outfit.
The number of likes follows a skewed distribution. We therefore normalise it with the logarithmic transform, namely $log(\text{likes} + 1)$ and scale it within the range of $(0, 1)$ with the use of min-max scaling. 
To create the time series used in QAR, we simply compute the weekly mean for each fashion category. The time series exhibit 10.72\% sparsity, so we apply linear interpolation to fill the missing values.
The dataset is split into 14,342,771 samples for \textit{established products} (training set) and 875,950 samples for \textit{new products} (validation and testing sets).

\subsubsection{\textbf{Mallzee}}
One limitation of SHIFT15m is that it provides no user-side information. Forecasting networks trained on SHIFT15m would learn to forecast the general popularity of a new garment design but the predictions would be identical regardless of age group or gender. For example, a newly designed ``floral dress'' would have the same popularity score for ``women, age: 18-25'' and ``men, age: 40-50''. 
In order to alleviate this issue and perform more targeted and useful forecasts for different segments of the market we collect the Mallzee dataset.

Project partner Mallzee collected 5,412,193 records from their databases between 14 demographic groups and 571,333 unique products. The selected demographics consist of two gender groups (men, women) and 7 age-groups including 0-18, 18-25, 25-30, 30-35, 35-45, 45-55, >55. The data span 1081 days between 2017-10-16 and 2020-09-30. We extract the visual features from a three-stage convolutional neural network consisting of an object detector, a garment category classifier and a fine-grained attribute detector, proposed in \cite{papadopoulos2022attentive}. We use the object-detector to identify individual garments in an image and the other two modules to extract the predicted fashion labels and the visual features from their last convolutional layers. The Mallzee dataset is classified into 22 garment categories such as blouses, dresses, jackets etc. and 109 fine-grained attributes including patterns (e.g. checked, quilted), prints (e.g. floral, graphic) and styles (e.g. bomber jacket, puffer jacket). The dataset is  split into 5,320,076 samples for \textit{established products} (training set) and 92,117 samples for \textit{new products} (validation and testing sets).

As the target variable, for each product $p$ we consider the popularity metric $P$ (Eq.\ref{eqn:popularity}) that expresses both \textit{likability} $L$ (Eq.\ref{eqn:likeability}) and \textit{reachability} $R$ (Eq.\ref{eqn:reachability}), given the set $A$ of the product's attributes, the target demographic group $G$ and the target time $t$:
\begin{subequations}
\begin{align}
\label{eqn:popularity}
P(p\mid A, G, t) & = L(p\mid G, t) \cdot R(A, G, s) \\ 
\label{eqn:likeability}
L(p\mid G, t) & = \frac{I(\oplus, G, p, t)}{I(\oplus, G, p, t) + I(\ominus, G, p, t)}\\
\label{eqn:reachability}
R(A, G, s) & = \prod_{a\in A}\frac{\mid \{u\in G\mid (u\leftrightarrow a)\land s\}\mid}{\mid \{u\in G\mid s\}\mid}
\end{align}
\end{subequations}
where $s$ is the year's season that day $t$ belongs to, $I(\oplus,G, p, t)$ and $I(\ominus,G, p, t)$ denote the number of positive and negative interactions between $G$ and $p$ at day $t$ respectively, $u$ denotes user, $\mid\cdot\mid$ denotes set cardinality and $u\leftrightarrow a$ denotes positive interaction of user $u$ with a product having attribute $a$. Hence, $likability$ is the probability that demographic group $G$ likes product $p$ at time $t$, while $reachability$ is the probability that demographic group $G$ interacts with the set of attributes $A$ at season $s$. The reason for incorporating reachability to calculate popularity is that positive interaction of small fractions of demographic groups with unusual attributes results in unexpected high likability. For instance, we found dresses, jumpsuits and heels in the top categories for men 18-25 based only on likability. The incorporation of reachability not only mitigated this issue but gave reasonable seasonal patterns to all garment categories as well. To create the time series used in QAR, we compute the weekly mean popularity for each of the 22 categories and 109 attributes which exhibit 1.13\% and 8.49\% sparsity respectively. We apply linear interpolation to fill the missing values. Time series of certain fashion classes can be seen in Figure \ref{fig:timeseries} where we can observe clear seasonal patterns. 

\subsubsection{\textbf{Amazon Reviews: Home and Kitchen}}

Fashion is the central focus of our work but we also deem important to examine the generalisability of MuQAR to other domains. 
To this end, we utilize the Amazon Reviews dataset\footnote{\url{https://jmcauley.ucsd.edu/data/amazon/}} and specifically the ``Home and Kitchen'' subset \cite{mcauley2015image}. 
Our selection criteria (cr.) for the dataset are: 
(1) relate to a primarily visually-driven domain, 
(2) provide a popularity metric, 
(3) provide images or extracted visual features,
(4) provide categories that produce dense time series.

We select the ``Home and Kitchen'' subset since it mostly contains products relating to furniture, decorative items, artwork posters and kitchen appliances.
We consider that visual appearances play a very important role in driving customer choices in this domain similarly to fashion (cr. 1).
We define the ‘star ratings’ as the target variable (cr. 2). 
The dataset provides the visual features extracted from the products' images by a convolutional neural network pre-trained on ImageNet (cr. 3).

The dataset spans from 1999 to 2014 and comes with 965 unique categories. However categories contain duplicates (e.g ``sheet'' and ``sheets'') and rare items (e.g ``charcoal drawings'' and ``crayon drawings'' only make a single appearance in the dataset). 
Aggregating the weekly time series for the 965 categories results in 64.34\% sparsity.
Even after filtering out the products before 2009 (a period with higher sparsity rates), results in 33.34\% sparsity. 
In order to mitigate this issue, we use the K-Means algorithm to cluster categories based on their TF-IDF textual representation into K=300 clusters. 
The weekly time series sparsity is reduced to 36.91\% for the whole dataset and to only 2.13\% for the filtered subset. (cr. 4). 
Filtering out the products before 2009 reduced the total size of the dataset from 3,261,846 to 3,002,786 which we consider a sensible trade-off between data loss and reduced time series sparsity. 
The dataset is split into 2,840,178 samples for \textit{established products} (training set) and 149,414 samples for \textit{new products} (validation and testing sets).

\subsection{Implementation Details}

We perform a grid search for hyper-parameter optimization for FusionMLP and QAR modules.
For FusionMLP we define $n_{mlp=}$3,4,5 or 6 fully connected layers with progressively narrower units $u_{mlp}=$ (2048, 1024, 512, 256, 128) (2048 units layer used twice in the case of 6 layers only) and an embedding space of $d_c=d_t=d_g=$8, 16 or 32 dimensions. 
For the QAR networks we define a hyper-parameter grid search as shown in Table \ref{tab:gridsearch}. For each QAR network we define the number of its component layers and their units. Values in parenthesis indicate experiments with progressively narrower units. 
After each LSTM, CNN or MLP layer we add a dropout layer of 10\% probability in order to reduce overfitting with the exception of Transformers which we define at 20\% probability.

For training we consider the Adam optimiser, cyclical learning rate \cite{smith2017cyclical} (initial learning rate: 1e-4, max learning rate: 1e-2, step size: 2, gamma: 0.1), the mean squared error (MSE) loss function for regression tasks (VISUELLE, SHIFT15m, Mallzee) and the categorical cross entropy for classification tasks (Amazon Reviews: Home and Kitchen).
We use a batch size of 1024 for Mallzee and Amazon datasets, 8192 for SHIFT15m since it is a larger dataset and we wanted to exploit parallelization and 16 for VISUELLE since it is a very small dataset.

\begin{table}
\centering
  \caption{Grid search for hyper-parameter optimization of the QAR networks in terms of the number of layers $n_{\{Layer\}}$ and their units $u_{\{Layer\}}$.}
  \label{tab:gridsearch}
  \begin{tabular}{c|c|c|c}
    \toprule
     \multicolumn{1}{c}{Model} & 
     \multicolumn{1}{c}{Layer} & 
     \multicolumn{1}{c}{$n_{\{Layer\}}$} & 
     \multicolumn{1}{c}{$u_{\{Layer\}}$}
     \\
     \midrule
     
     LSTM & lstm & 1, 2, 3 & (512, 256, 128) \\
     \midrule
     CNN & cnn & 1, 2, 3 & (512, 256, 128) \\
     \midrule
     
     DA-RNN & lstm & 2 & 64 or 128 \\
     \midrule
     
      \multirow{2}{*}{ConvLSTM} & cnn & 1, 2, 3 & (512, 256, 128) \\
      & lstm & 1, 2, 3 & (512, 256, 128) \\
    \midrule
    
    \multirow{2}{*}{F-LSTM} & lstm & 1, 2, 3 & (512, 256, 128) \\
    & mlp & 0,1 & 256 or 512 \\
     \midrule
     
    \multirow{2}{*}{Transformer} & block & 1, 2, 3 & \multirow{2}{*}{128, 256 or 512} \\
    & head & 2 or 4 &  \\

    \bottomrule
  \end{tabular}
  
\end{table}

\begin{table*}
% \scriptsize
\footnotesize
\centering
  \caption{Popularity forecasting models trained on `established' and evaluated on `new' garments for three datasets: Mallzee (MLZ), SHIFT15m and Amazon Reviews: Home and Kitchen.
  Features used: Images [I], target attributes time series [A], exogenous attributes time series [X] and image captions [C].
  [A] and [X] have 12 weeks-long time series as input. 
  The models forecast the next week. \\
  \textbf{Bold} denotes the best overall performance per metric and dataset. 
  \underline{Underline} denotes the best performing QAR network per dataset; which are used in the final MuQAR models. \\
  *\textit{[C] are only available on MLZ; they are ignored on Amazon and SHIFT15m, [I+A+X] are used instead.}
  }
  \label{tab:results_ablation}
  \begin{tabular}{cc|cc|cc|ccc|c}
    \toprule
     \multicolumn{1}{c}{\textbf{Input}} & 
     \multicolumn{1}{c}{\textbf{Model}} & 
     \multicolumn{2}{c}{\textbf{MAE($\downarrow$) }} & 
     \multicolumn{2}{c}{\textbf{PCC($\uparrow$) }} & 
     \multicolumn{3}{c}{\textbf{Accuracy($\uparrow$) }} &
     \multicolumn{1}{c}{\textbf{AUC($\uparrow$) }} 
     \\
     
    \midrule
    
    && MLZ & SHIFT15m & MLZ & SHIFT15m & MLZ & SHIFT15m & Amazon & Amazon\\
    
    \midrule
    
    \multirow{6}{*}{[A]} &
    LR & 0.1878 & 0.1162 & 0.2439 & 0.3177 & 63.10 & 59.58 & 48.52 & 65.41 \\
    
    &
    CNN & \underline{0.1611} & 0.1148 & \underline{0.5379} & 0.3406 & \underline{70.99} & 61.51 & 47.18 & 69.34 \\
    
    &
    LSTM & 0.1656 & 0.1150 & 0.5109 & 0.3371 & 69.67 & 61.42 & 45.58 & 67.54 \\
    
    &
    F-LSTM & 0.1809 & 0.1149 & 0.3395 & 0.3376 & 64.62 & 61.43 & 44.95 & 67.89 \\
    
    &
    Transformer & 0.1842 & 0.1149 & 0.3071 & 0.3398 & 63.67 & 61.28 & \underline{51.10} & \underline{71.29} \\
    
    &
    ConvLSTM & 0.1641 & \underline{0.1147} & 0.5225 & \underline{0.3411} & 69.98 & \underline{61.58} & 46.58 & 68.59 \\
    
    \midrule
    
    \multirow{2}{*}{[A+X]} & ConvLSTM+X & \underline{0.1686} & \underline{0.1185} & \underline{0.4913} & \underline{0.2191} & \underline{68.86} & 59.50 & 60.61 & \underline{78.95} \\
    & DA-RNN & 0.1863 & 0.1187 & 0.2652 & 0.2050 & 64.39 & \underline{59.55} & 60.61 & 78.90 \\
    
    \midrule
    
    \multirow{2}{*}{[I]} &
    LR & 0.1599 & 0.1186 & 0.5314 & 0.1940 & 71.86 & 57.93 & 41.46 & 68.18 \\
    
    &
    FusionMLP & 0.1074 & 0.1148 & 0.7893 & 0.2811 & 81.52 & 60.89 & 46.96 & 71.69 \\
    
    \midrule
    
    [I+C] & FusionMLP & 0.1073 & - & 0.7879 & - & 81.30 & - & - & - \\
        
    \midrule
    
    [I+A] &
    MuQAR & 0.0949 & \textbf{0.1100} & 0.8362 & \textbf{0.3934} & 83.41 & \textbf{63.57} & 51.51 & 74.24 \\
    
    \midrule
    
    [I+A+X+C] &
    MuQAR & \textbf{0.0911} & 0.1118* & \textbf{0.8484} & 0.3448* & \textbf{84.26} & 62.30* & \textbf{60.63*} & \textbf{80.40*} \\

    \bottomrule
  \end{tabular}
\end{table*}

\begin{table*}
\centering
  \caption{Comparative analysis between MuQAR and its modules against state of the art networks on the VISUELLE dataset using 52 week-long time series as input from Google Trends and forecasting the next 6. Features used: [T]ext, [I]mage, target [A]ttribute time series from Google trends, e[X]ogenous time series from google trends and image [C]aptions.
  \underline{Underline} denotes the best performing network per input type. \textbf{Bold} denotes the best overall performance.
  }
  \label{tab:results_visuelle}
  \begin{tabular}{l|c|cc}
    \toprule
     \multicolumn{1}{c}{\textbf{Method}} & 
     \multicolumn{1}{c}{\textbf{Input}} & 
     \multicolumn{2}{c}{\textbf{IN:52, OUT:6}}
     \\
     
     \midrule
     & & WAPE($\downarrow$) & MAE($\downarrow$) \\
     
    \midrule
    GTM-Transformer \cite{skenderi2021well} & \multirow{3}{*}{[T]} & 62.6 & 34.2 \\    
    Attribute KNN \cite{ekambaram2020attention} &  & 59.8 & 32.7 \\
    FusionMLP &  & \underline{55.15} & \underline{30.12} \\
    \midrule
    
    Image KNN \cite{ekambaram2020attention} & \multirow{3}{*}{[I]} & 62.2 & 34 \\
    GTM-Transformer \cite{skenderi2021well} &  & 56.4 & 30.8 \\
    FusionMLP &  & \underline{54.59} & \underline{29.82} \\
    
    \midrule

    Transformer & \multirow{6}{*}{[A]} & 62.5 & 34.1 \\
    LSTM &  & 58.7 & 32.0 \\
    ConvLSTM &  & 58.6 & 32.0 \\    
    
    GTM-Transformer \cite{skenderi2021well} &  & 58.2 & 31.8 \\
    
    F-LSTM & & 58.0 & 31.7 \\
    CNN &  & \underline{57.4} & \underline{31.4} \\
    
    \midrule
    
    ConvLSTM+X & \multirow{2}{*}{[A + X]} & \underline{55.73} & \underline{30.44} \\
    DA-RNN &  & 58.05 & 31.71 \\
        
    \midrule
    
    Attribute + Image KNN  \cite{ekambaram2020attention} & \multirow{4}{*}{[T + I]} & 61.3 & 33.5 \\
    Cross-Attention RNN \cite{ekambaram2020attention} &  & 59.5 & 32.3 \\
    GTM-Transformer \cite{skenderi2021well} &  & 56.7 & 30.9 \\    
    FusionMLP &  & \underline{54.11} & \underline{29.56} \\    
    \midrule
    
    FusionMLP & [T+I+C] & 53.50 & 29.22 \\    
    
    \midrule

    GTM-Transformer AR \cite{skenderi2021well} & \multirow{8}{*}{[T + I + A]} & 59.6 & 32.5 \\    
    
    Cross-Attention RNN+A \cite{ekambaram2020attention} & & 59.0 & 32.1\\
    
    GTM-Transformer \cite{skenderi2021well} & & 55.2 & 30.2 \\    

    MuQAR w/ Transformer & & 54.87 & 29.97\\
    MuQAR w/ F-LSTM & & 54.37 & 29.7 \\    
    MuQAR w/LSTM & & 54.3 & 29.66 \\

    MuQAR w/CNN & & 53.9 & 29.44 \\
    MuQAR w/ ConvLSTM &  & \underline{53.61} & \underline{29.28} \\
    
    \midrule
    MuQAR w/ DA-RNN & \multirow{2}{*}{[T+I+A+X+C]} & 54.43 & 29.73 \\    
    
    MuQAR w/ ConvLSTM+X & & \underline{\textbf{52.63}} & \underline{\textbf{28.75}} \\    
    
    \bottomrule
  \end{tabular}
  
\end{table*}

\section{Results}
\label{section:results}

\subsection{Ablation Analysis}
\label{sec:ablation}

Table \ref{tab:results_ablation} presents the ablation analysis of MuQAR on Mallzee, SHIFT15m and Amazon Reviews datasets. 

\textbf{QAR}: By performing the hyper parameter optimisation grid search on the Mallzee dataset we identify the following best performing combinations, LSTM: $u_{lstm}=(512)$ units, CNN: $u_{cnn}=(512, 256, 128)$, DA-RNN: $u_{lstm}=(128)$, ConvLSTM: $u_{cnn}=(512, 256, 128)$, $u_{lstm}=(512, 256)$ and $u_{mlp}=(128)$, F-LSTM: $u_{lstm}=(256)$ and $u_{mlp}=(512)$, Transformer: $n_{block}=1$, $n_{head}=2$ with 256 units, ConvLSTM+X: we use the same hyper parameters as ConvLSTM.
We apply the same hyper-parameters on the rest of the datasets.
The CNN, ConvLSTM and Transformer QAR networks yielded the best performance for Mallzee, SHIFT15m and Amazon Reviews datasets respectively when only using [A], the target attribute time series with 12 previous weeks as input and 1 as output. 
We integrate these specific QAR models for each of the three datasets in the MuQAR experiments. We can observe that there is not a single QAR network that consistently performs better. Experimentation is required in order to identify the most appropriate QAR network for specific tasks and datasets.

When QAR utilizes both [A + X] its performance decreases on Mallzee and SHIFT15m; compared to the best QAR with [A].
However, we observe an impressive improvement on the Amazon dataset with +18\% and +10\% increases in terms of Accuracy and AUC respectively. 
This may be attributed to the fact that Amazon initially had 965 categories which we clustered into 300 with the use of K-Means - a lot more than in Mallzee and SHIFT15m - some of which may be quite noisy. Thus, feeding the exogenous time series [X] may help QAR discern the informative from the noisy time series.
Overall, we observe that ConvLSTM+X tends to outperform DA-RNN in 6 out of 8 cases. We therefore integrate ConvLSTM+X in the final MuQAR experiments.

\textbf{FusionMLP}: an embedding space of $d_c=d_t=d_g$ = 32 dimensions, $n_{mlp}=4$ with $u_{mlp}=(2048, 1024, 512, 256)$ and a dropout rate of 10\% performed best on the three datasets.
FusionMLP performed significantly better than the QAR networks on the Mallzee dataset but not as good on SHIFT15m and Amazon Reviews.
The visual features [I] in the Mallzee data are extracted by a fined-tuned network on fashion imagery, while the other two datasets utilized networks pre-trained on ImageNet and are therefore less specialised. This fact may have affected the results since, presumably, the quality and specialisation of the extracted visual features plays a crucial role in the task.
When image captions [C] are added in FusionMLP there is only a negligible improvement in terms of MAE but a small decrease in terms of PCC and Accuracy. This issue may be due to the hyper-parameter tuning being done on the [I] features and not fine-tuned for [I+C].

\textbf{MuQAR}: we can observe that MuQAR using [I+A] is capable of consistently surpassing all QAR networks and FusionMLP on the three datasets.
By combining the visual features [I] with the target time series [A], MuQAR is able to improve upon the task of forecasting the popularity of new products while being robust to less specialised visual features. 
Finally, utilizing exogenous time series [X] and captions [C] further improves the MuQAR architecture for Mallzee and Amazon datasets but not on SHIFT15m. 
While [A+X] did not consistently improve the performance of QAR, neither did [C] improve FusionMLP, when integrated within the MuQAR architecture they can show significant improvements.

\begin{figure*}[]
  \begin{subfigure}[b]{0.5\textwidth}
    \includegraphics[width=1\textwidth]{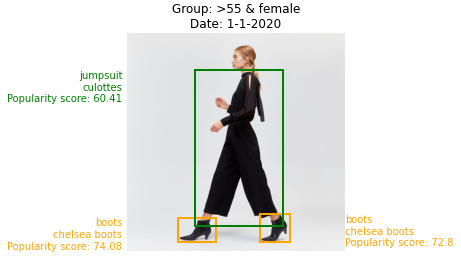}
    \caption{}
    \label{fig:f1}
  \end{subfigure}
  \hfill
  \begin{subfigure}[b]{0.5\textwidth}
    \includegraphics[width=1\textwidth]{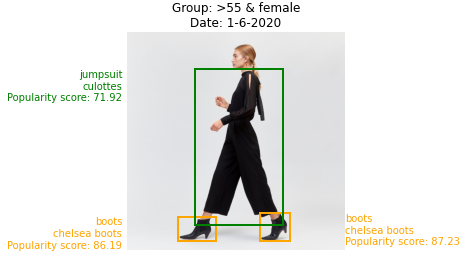}
    \caption{}
    \label{fig:f2}
  \end{subfigure}
  \hfill
  \begin{subfigure}[b]{0.5\textwidth}
    \includegraphics[width=1\textwidth]{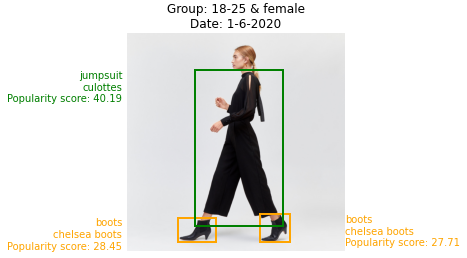}
    \caption{}
    \label{fig:f3}
  \end{subfigure}
  \hfill  
  \begin{subfigure}[b]{0.5\textwidth}
    \includegraphics[width=1\textwidth]{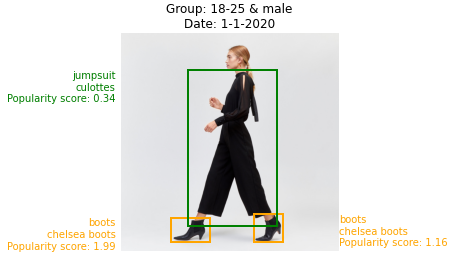}
    \caption{}
    \label{fig:f4}
  \end{subfigure}
  \hfill  
  \begin{subfigure}[b]{0.5\textwidth}
    \includegraphics[width=0.7\textwidth]{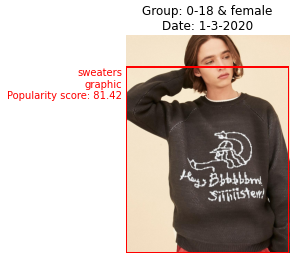}
    \caption{}
    \label{fig:f5}
  \end{subfigure}
  \hfill  
  \begin{subfigure}[b]{0.5\textwidth}
    \includegraphics[width=0.7\textwidth]{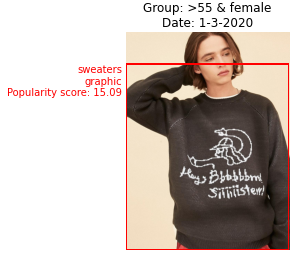}
    \caption{}
    \label{fig:f6}
  \end{subfigure} 
    \begin{subfigure}[b]{0.5\textwidth}
    \includegraphics[width=0.7\textwidth]{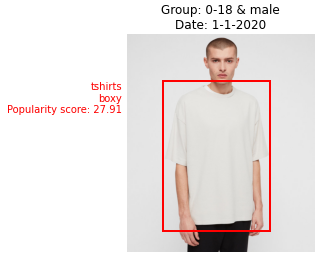}
    \caption{}
    \label{fig:f7}
  \end{subfigure}
  \hfill
  \begin{subfigure}[b]{0.5\textwidth}
    \includegraphics[width=0.7\textwidth]{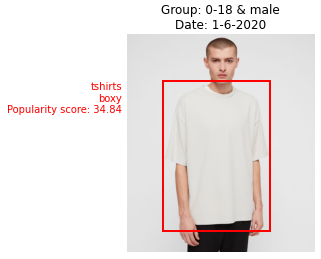}
    \caption{}
    \label{fig:f8}
  \end{subfigure}
\caption{Inference samples by the MuQAR on Mallzee data for different demographic groups and dates.}  
\label{fig:inference}
\end{figure*}

In Figure \ref{fig:inference} we present an inference sample predicted by MuQAR trained on the Mallzee dataset. We use the same image, depicting a jumpsuit and a pair of chelsea boots and MuQAR performs predictions for different demographic groups. We can observe that the popularity of all garments increases in June for women over 55 compared with January (Fig. \ref{fig:f1} and \ref{fig:f2}). Moreover, the outfit is less popular with younger women (Fig. \ref{fig:f3}) and very unpopular with men (Fig. \ref{fig:f4}). 
In Figures \ref{fig:f5} and \ref{fig:f6} we observe that a sweater with a ``playful'' graphic design is more popular with younger (mostly teenagers) than older women.
Finally, a minimal white ``boxy t-shirt'' shows an increase in popularity from January (Fig. \ref{fig:f7}) to June (Fig. \ref{fig:f7}) presumably due to the seasonal change and by extension the warmer weather - while still remaining relatively unpopular for young teenage boys.
MuQAR seems to have learned both seasonal trends and the average preferences of different demographic groups. 

\subsection{Comparative Analysis}
\label{section:comparative}

Table \ref{tab:results_visuelle} presents the comparison between MuQAR and its modules against state of the art networks proposed by \cite{ekambaram2020attention} and \cite{skenderi2021well} on the VISUELLE dataset. 
We do not fine tune the hyperparameters of MuQAR and its modules on the VISUELLE dataset. Rather, we use the same values as described in \ref{sec:ablation} with the exception of increasing the dropout rate to 0.3 to avoid overfitting since VISUELLE is a significantly smaller dataset. 

Regarding QAR using [A], CNN and F-LSTM were able to surpass both GTM-Transformer and our Transformer. This indicates that it is advisable to also experiment with ``simpler'' neural network architectures and not immediately leaping to more complicated networks. Moreover, ConvLSTM+X utilizing both target and exogenous time series [A+X] surpasses all QAR models using [A]. 
FusionMLP utilizing [T+I], or solely [T] or [I] can not only outperform its similar-input competitors but also the GTM-Transformer using [T+I+A].
On top of that, adding image captions [C] improves the performance of FusionMLP. 
We also observe that MuQAR with any of the QAR networks is capable of surpassing GTM-Transformer (non-AR), Cross Attention RNN+A and GTM-Transformer AR (an autoregressive variant) when using [T+I+A]. Finally, the proposed MuQAR w/ ConvLSTM+X employing all features [T+I+A+X+C] is capable of surpassing all other models, setting a new state-of-the-art on VISUELLE with 4.65\% and 4.8\% improvements in terms of WAPE and MAE respectively. These results further prove the validity of our two main proposals, the use of image captioning and the inclusion of exogenous attributes time series within the MuQAR architecture when forecasting the visual popularity of new garment designs.

\section{Conclusions}
\label{section:conclusion}

In this study we propose MuQAR, a Multimodal Quasi Auto-Regressive deep learning architecture, for forecasting the popularity of new products that lack historical data.
The proposed architecture consists of two modules: (1) a multi-modal multilayer perceptron (FusionMLP) representing visual, textual, categorical and temporal aspects of a product and (2) a quasi-autoregressive (QAR) neural network modelling the time series of the product's attributes along with all other attributes time series. 
For FusionMLP, we extract the visual and categorical features (fashion attributes) from a computer vision model representing the unique characteristics of new products along with rich textual descriptions extracted from an image captioning model. 
In QAR, the time series of its attributes provide a proxy of temporal patterns for the lack of historical data while the exogenous time series are used to identify relations among the target and all other attributes.

Our focus is centered around the fashion industry and new garment designs. We experiment with three fashion image datasets: Mallzee, SHIFT15m and VISUELLE. We also experiment with the Amazon Reviews: Home and Kitchen dataset to examine the generalisability of the proposed architecture. 
Performing an extensive internal ablation analysis as well as a comparative analysis, we show that MuQAR is capable of competing and surpassing the domain's current state of the art by 4.65\% in terms of WAPE and 4.8\% in terms of MAE on the VISUELLE dataset.

In this study, we experiment with two QAR models that utilize the exogenous time series, namely DA-RNN and ConvLSTM+X. It would be interesting for future research to examine how other models perform within the QAR module such as Lavarnet \cite{koutlis2020lavarnet}, MTNet \cite{chang2018memory}, LSTNet \cite{lai2018modeling} or the Informer \cite{zhou2021informer}.
Moreover, we have only experimented with image datasets but the proposed architecture MuQAR could be adapted and applied to other domains with different types of data. For example audio feature combined with time series of musical genres could be used to forecast the popularity of new tracks or albums. 
In future work we plan on examining how visual-temporal features extracted from MuQAR can facilitate recommendation systems and especially in mitigating the cold start problem of new items.

\section{Acknowledgments}
This work is partially funded by the project ``eTryOn - virtual try-ons of garments enabling novel human fashion interactions'' under grant agreement no. 951908. The authors would also like to thank Jamie Sutherland, Manjunath Sudheer and the company Mallzee/This Is Unfolded for the data acquisition as well as all the useful insights and feedback.

\bibliographystyle{unsrt}  
\bibliography{references}

\end{document}